\pdfoutput=1

\documentclass[11pt]{article}

\usepackage[final]{acl}

\usepackage{times}
\usepackage{latexsym}

\usepackage[T1]{fontenc}

\usepackage[utf8]{inputenc}

\usepackage{microtype}

\usepackage{inconsolata}

\usepackage{graphicx}

\usepackage{amsmath}
\usepackage{amssymb}
\usepackage[ruled,vlined,algo2e,linesnumbered]{algorithm2e}

\title{MaskPrune: Mask-based LLM Pruning for Layer-wise Uniform Structures}

\author{
 \textbf{Jiayu Qin\textsuperscript{1,2,*}},
 \textbf{Jianchao Tan\textsuperscript{2,*}},
 \textbf{Kefeng Zhang\textsuperscript{2}},
 \textbf{Xunliang Cai\textsuperscript{2}},
 \textbf{Wei Wang\textsuperscript{1,+}}
\\
\\
 \textsuperscript{1}Nanjing University,
 \textsuperscript{2}Meituan
\\
 \small{
   \textbf{Correspondence:} jiayuqin@smail.nju.edu.cn, \{tanjianchao02, zhangkefeng, caixunliang\}@meituan.com, ww@nju.edu.cn
 }
}

\begin{document}
\maketitle
\begin{abstract}
The remarkable performance of large language models (LLMs) in various language tasks has attracted considerable attention. However, the ever-increasing size of these models presents growing challenges for deployment and inference. Structured pruning, an effective model compression technique, is gaining increasing attention due to its ability to enhance inference efficiency. Nevertheless, most previous optimization-based structured pruning methods sacrifice the uniform structure across layers for greater flexibility to maintain performance. The heterogeneous structure hinders the effective utilization of off-the-shelf inference acceleration techniques and impedes efficient configuration for continued training. To address this issue, we propose a novel masking learning paradigm based on minimax optimization to obtain the uniform pruned structure by optimizing the masks under sparsity regularization. Extensive experimental results demonstrate that our method can maintain high performance while ensuring the uniformity of the pruned model structure, thereby outperforming existing SOTA methods.
\end{abstract}


\section{Introduction}

Large Language Models (LLMs), such as OpenAI's GPT series \citep{achiam2023gpt} and Meta's LLaMA \citep{touvron2023llama, touvron2023llama2}, have made substantial advancements in the domain of Natural Language Processing (NLP). These models exhibit robust capabilities in language understanding and generation, facilitated by extensive pre-training and fine-tuning. However, as the size of these models continues to expand, their computational and storage demands increase sharply, presenting significant challenges for practical applications. Model compression, a vital approach to reducing memory footprint and computational load during model deployment, offers unique benefits across various domains. Techniques such as pruning \citep{frantar2023sparsegpt, ma2023llm, sun2023wanda}, quantization \citep{frantar2023gptq, xiao2023smoothquant, lin2024awq}, knowledge distillation \citep{gu2024minillm, agarwal2023gkd}, and low-rank factorization \citep{yuan2023asvd, wang2024svd} can significantly decrease the number of model parameters and computational complexity, thereby enabling large-scale language models to function efficiently in resource-constrained environments.

\begin{figure}
    \centering
    \includegraphics[width=1\linewidth]{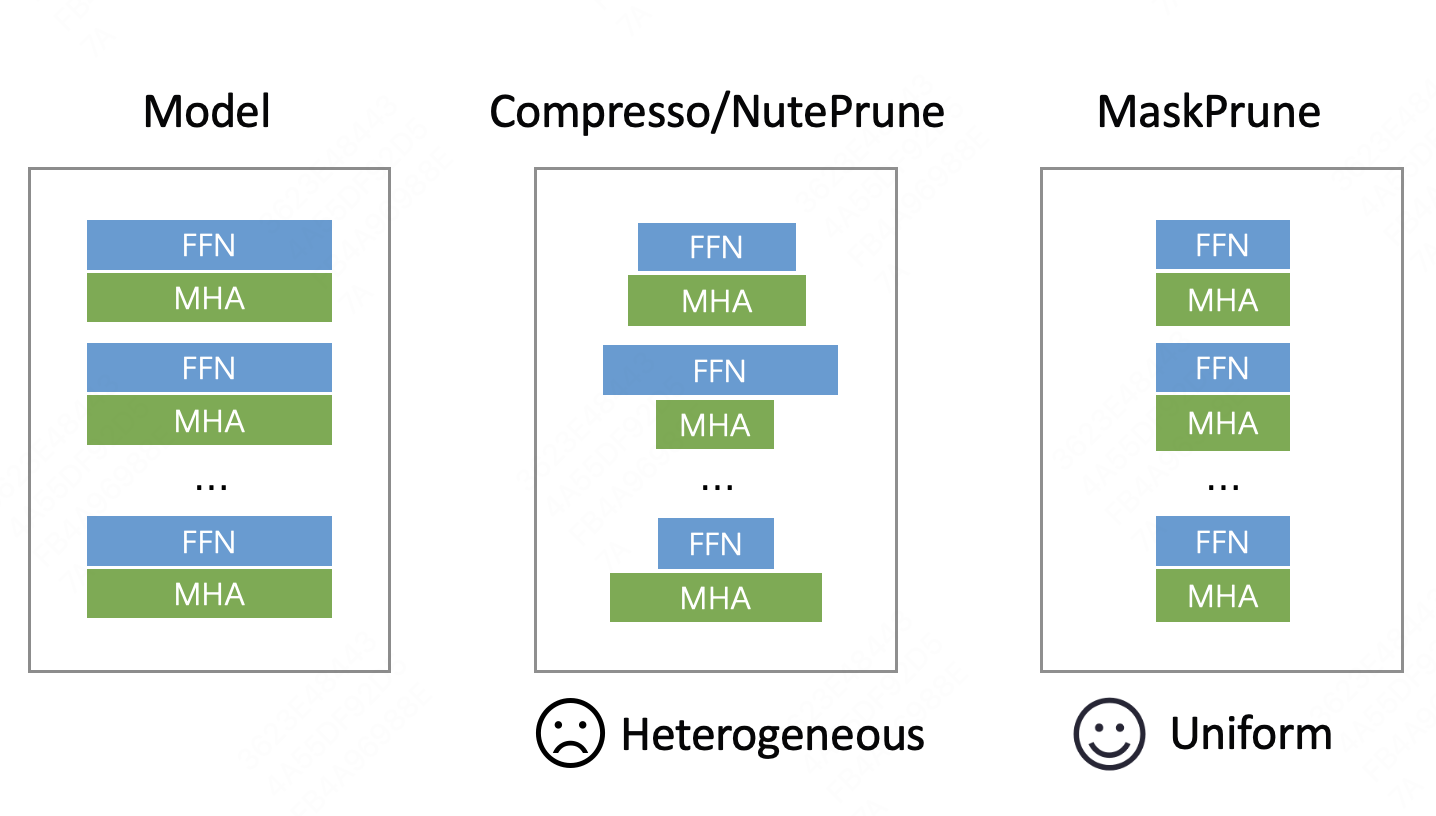}
    \caption{Compresso/NutePrune results in heterogeneous inter-layer structures, whereas MaskPrune achieves uniform inter-layer structures, which is friendly to inference deployment and continual training.}
    \label{fig:MaskPrune1}
\end{figure}

The pruning technique reduces the size and computational complexity of the models by eliminating redundant parameters, which can generally be categorized into unstructured pruning \citep{frantar2023sparsegpt, sun2023wanda, dong2024pruner}, semi-structured pruning \citep{mishra2021accelerating}, and structured pruning \citep{ma2023llm, xia2023sheared, an2023fluctuationbased}. Unstructured pruning compresses models by removing individual parameters, resulting in sparse weight matrices that consume less memory. However, without dedicated hardware support, the updated models do not achieve faster inference, thereby still imposing computational burdens during the inference process. Semi-structured pruning offers some speed improvements, but these are limited compared to those achieved by structured pruning. Structured pruning adopts a more modular approach to remove modules from models, typically targeting attention heads, embedding dimensions, FFN intermediate dimensions, experts in Mixture-of-Experts (MoE) networks, or layers. After structured pruning, the weight matrices of the models remain dense, and their reduced dimensions typically lead to greater inference acceleration. However, the coarser granularity of this pruning method makes it more challenging to preserve model capabilities after pruning. Currently, most pruning techniques employ metric-based methods, which determine the modules to be pruned by introducing specific pruning metrics. These metrics are usually designed heuristically and often perform poorly at high pruning rates. Moreover, a single metric cannot fully capture the importance of model weights, making it difficult to identify superior local optimal solutions. In contrast, optimization-based pruning methods determine which weights to prune by learning a pruning mask, thereby avoiding the performance degradation associated with manually designed metrics. This paper primarily focuses on optimization-based pruning methods.

Given the large scale of Large Language Models (LLMs), existing optimization-based pruning methods employ structured pruning, wherein a single mask prunes entire modules of the model. Methods such as CoFi \citep{xia2022structured}, Compresso \citep{guo2023compresso} and NutePrune \citep{li2024nuteprune} follow the $L_0$ regularization \citep{louizos2018learning} training paradigm during the training of pruning masks, learning masks by setting a total sparsity without additional constraints. This approach results in a lack of uniformity between layers during training, causing each layer to have a different number of attention heads and FFN intermediate dimensions, as illustrated in Figure \ref{fig:MaskPrune1}, which leads to suboptimal inference speed. Moreover, this irregular structure necessitates adaptations during model deployment. Additionally, to achieve higher performance post-compression, existing model compression techniques typically involve continued training and fine-tuning after compression. However, the irregular structure hinders models from fully utilizing existing model parallelism techniques, resulting in diminished performance for the same continued training cost\citep{xia2023sheared}.

To address these issues, this paper proposes a method called MaskPrune for jointly training pruning masks and target structures across various dimensions. This approach optimizes the target dimension parameters simultaneously during training to maintain uniformity of dimensions across the layers of the pruned model while achieving the preset model sparsity. The key idea is to frame the sparsity constraint of model pruning as a minimax problem. Since the introduced sparsity loss is non-differentiable, it cannot be directly optimized using gradient descent. By employing proximal operators and straight-through estimators to optimize masks and target dimensions respectively, the pruning optimization problem is effectively solved. The contributions of this paper can be summarized as follows:

\begin{itemize} 
    \item We propose a mask training method based on minimax optimization, enabling end-to-end optimization of mask values during the pruning and automatically maintaining the layerwise uniform structure throughout training. 
    \item Mask parameters are optimized by proximal operators, maintaining the original model's capabilities to the greatest extent while adhering to target sparsity constraints and minimizing performance degradation during pruning.
    \item Extensive experiments were conducted across various sparsity levels on models from the LLaMA family, demonstrating the effectiveness of our method by maintaining high performance on diverse tasks while preserving the model's uniform structure.
\end{itemize}

\section{Related Work}

\paragraph{Importance metric-based Methods}SparseGPT \cite{frantar2023sparsegpt} evaluates the importance of weights using second-order Hessian information and compensates for other weights during the pruning process, thereby achieving unstructured pruning. Wanda \cite{sun2023wanda} simplifies this approach by relying solely on the magnitude of the weights and the activation values on a calibration set to determine the importance of the weight, accelerating the pruning process. Additionally, its methods can be extended to semi-structured pruning. Pruner-Zero \cite{dong2024pruner} employs genetic programming to efficiently search for optimal symbolic pruning metrics, avoiding heuristic weight importance searches. LLM-Pruner \cite{ma2023llm} was the first to utilize structured pruning methods to compress large language models (LLMs), assessing the importance of weight groups through approximate first-order Hessian information. Bonsai \cite{dery2024everybody} samples the correlation between sub-modules and model performance, using linear regression to determine the importance of weight groups. LoRAPrune \cite{zhang2023loraprune} estimates the original gradients of weights through the gradients of LoRA matrices, thereby reducing memory consumption during backpropagation.

\paragraph{Optimization-based Methods}However, metric-based methods like LLM-Pruner \cite{ma2023llm} often fail to fully capture the importance of weights, leading to suboptimal generalization performance. To address this, many optimization-based methods have focused on learning masks for pruned weights. $L_0$ regularization \cite{louizos2018learning} offers a general paradigm for mask learning, enabling the optimization of non-differentiable masks. CoFi \cite{xia2022structured} integrates a hierarchical distillation loss into the training loss function, while SheardLlama \cite{xia2023sheared} specifies target structures to achieve a unified model architecture and employs dynamic batch loading to enhance generalization performance. Compresso \cite{guo2023compresso} introduces specific prompts during training and incorporates LoRA modules into the optimization process, combining fine-tuning with mask training. NutePrune \cite{li2024nuteprune} leverages progressive distillation to enhance the transfer of knowledge from teacher to student models.
\section{Methodology}
\begin{figure}
    \centering
    \includegraphics[width=1\linewidth]{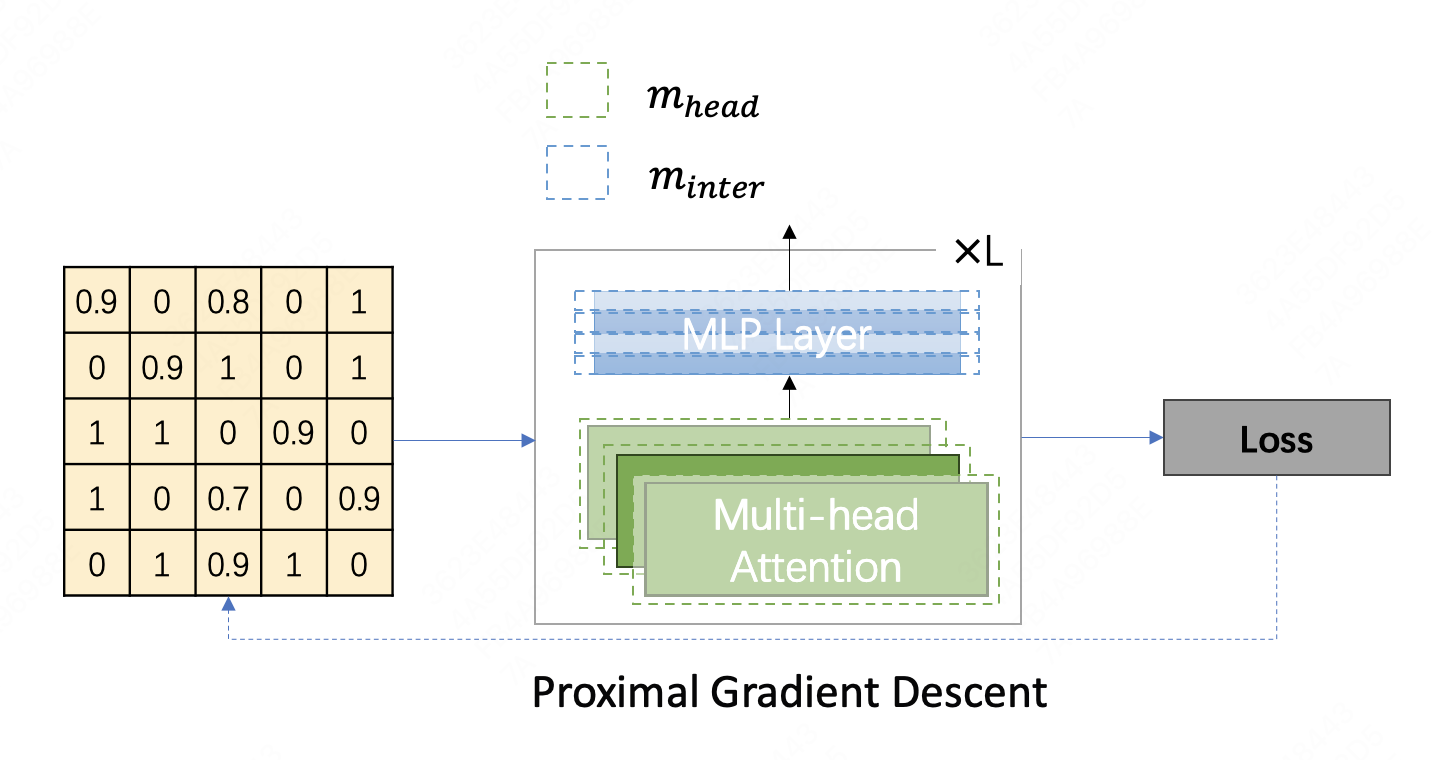}
    \caption{Overall framework of MaskPrune. We optimize mask values through proximal gradient updates to identify the optimal pruning structure while simultaneously fine-tuning other parameters.}
    \label{fig:MaskPrune2}
\end{figure}

In this chapter, we explain how MaskPrune employs an optimization-based approach to generate structured pruning masks while maintaining consistency across inter-layer structures throughout the process. Specifically, Section \ref{Problem Definition} defines the optimization problem and Section \ref{Parameter Update Strategy} introduces the methods to solve this problem. Figure \ref{fig:MaskPrune2} illustrates the overall framework of our proposed method.

\subsection{Problem Definition} \label{Problem Definition}

For the transformer models such as Llama families, the Multi-Head Attention (MHA) layer and the Feed-Forward Network (FFN) comprise the primary components. During the pruning, the main structural elements, specifically, the attention heads and the intermediate dimensions of the FFN, are typically processed. To facilitate pruning, the corresponding masks, $m_{head}$ and $m_{inter}$ are introduced at their respective positions within the model. Taking the Llama architecture as an example:

\begin{equation*}
MHA^{l}(X) = \sum_{j=1}^{N_{h}} m_{head}^{(l, j)} \cdot \text{Attn}^{(l, j)}(X)
\end{equation*}

where \( l \) denotes the \( l \)-th layer and \( N_h \) denotes the number of attention heads in the Multi-Head Attention layer.

{\small
\begin{equation*}
FFN^{l}(X) = m^l_{inter} \cdot \left( W^l_{gate}(X) \odot W^l_{up}(X) \right) \cdot W^l_{down}(X)
\end{equation*}
}

Here, $m_{head}$ and $m_{inter}$ are restricted within the range $[0,1]$. When a mask value is 0, the corresponding module is pruned. The actual sparsity of the model is calculated as follows:

\begin{equation*}
\begin{aligned} 
\hat{s} &= \frac{1}{M} \cdot 4 \cdot d_{h} \cdot d_{hidden} \cdot \sum_{l=1}^{L} \sum_{j=1}^{N_{h}} \mathbb{I}\left(m_{head}^{(l, j)}=0\right) \\
&\quad + \frac{1}{M} \cdot 3 \cdot d_{hidden} \cdot \sum_{l=1}^{L} \sum_{j=1}^{d_{f}} \mathbb{I}\left(m_{inter}^{(l, j)}=0 \right)
\end{aligned}
\end{equation*}

Here, \( M \) denotes the original size of the model, \( L \) is the total number of layers in the model, \( d_{hidden} \) represents the hidden dimension, and \( d_f \) and \( d_h \) signify the head dimension and the intermediate dimension within the FFN, respectively. The indicator functions are used to identify the pruned components of the model, these components constitute the pruned model. The objective of our method is to implement structured pruning using a regularized approach. Therefore, the optimization target encompasses not only the pruning mask $m$ but also the sparsity parameters $s=\{s_{head},s_{inter}\}$ across various dimensions, where $s_{head}$ and $s_{inter}$ represent the sparsity ratios of the MHA layers and FFN layers, respectively. For clarity, $s$ below denotes the number of pruned dimensions, which is the product of the original dimensions and the sparsity.

The sparse constraint problem for the model can be reformulated as follows \cite{tono2017efficientdcalgorithmconstrained}, where \( m \in \{ m_{head}, m_{inter} \} \) and \( s \in \{ s_{head}, s_{inter} \} \):

\begin{equation*}
\sum_j \mathbb{I} \left(  m^{(l,j)} = 0 \right) \geq s \iff \left\| m^{l} \right\|_{s,2} = 0
\end{equation*}

where \( \left\| m^{l} \right\|_{s,2} \) represents the 	$\ell_2$ norm of a subvector of \( m^{l} \), consisting of the \( s \) elements with the smallest norms. This constraint ensures that, for a given dimension's mask, the $s$ smallest elements in each layer's mask are zero, meaning $s$ dimensions are pruned, thereby maintaining the uniformity of the mask structure across layers.

Our method aims to compress the model under resource constraints. Given a resource target, the algorithm will continuously compress the model until the target is met. In the context of model compression, the resource constraint typically refers to the model's memory footprint. When the target size of the model after pruning is ${M}_{prune}$, it satisfies

\begin{equation*}
M(s) \leq M_{prune}
\end{equation*}

where \( {M}(s) \) quantifies the memory footprint of the model after pruning based on sparsity \( s \):
{
\small
\begin{equation*}
M(s) = M - L \cdot (4 \cdot d_{hidden} \cdot d_{h} \cdot s_{head} + 3 \cdot d_{hidden} \cdot s_{inter})
\end{equation*}
}

In summary, our final optimization goal can be described as a minimax problem:


\begin{align*} 
&\min_{\{m,s\}}\max_{\{y,z\geq0\}}\mathcal{L}_{\mathrm{pruning}} = \min_{\{m,s\}}\max_{\{y,z\geq0\}}\mathcal{L}(m) \\ & + \underbrace{y \sum_{l=1}^{L} \left( \left\|m_{head}^{l}\right\|_{\lceil s_{head} \rceil,2}^2 + \left\|m_{inter}^{l}\right\|_{\lceil s_{inter} \rceil,2}^2 \right)}_{\mathrm{sparsity~loss}} \\ & + \underbrace{z\left(M(s)-M_{prune}\right)}_{\mathrm{resource~loss}} 
\end{align*}

Ultimately, we utilize the solution $s$ to the aforementioned minimax problem to determine the compression ratio for each layer. In this process, we directly sort the norms of the elements, selecting and removing those with the smallest norms to perform pruning.

\subsection{Parameter Update Strategy} \label{Parameter Update Strategy}

We iteratively solve the aforementioned optimization problem. Following the approach of \cite{tono2017efficientdcalgorithmconstrained, Chen2023ResourceCM, yu2022unified}, we first update the mask to optimize the training loss. Subsequently, we update \( s \), which balances the sparsity loss and resource loss, ultimately achieving a convergence state. Finally, we update the variables \( y \) and \( z \) to increase the penalties associated with the sparsity and resource losses, thereby promoting the convergence of \( s \). The specific update strategy is as follows:

In this optimization step, we fix the model parameters, the sparsity variable \( s \), and the Lagrange multipliers \( y \) and \( z \) while updating \( m \). Following the methodology of \cite{yang2019ecc}, we minimize the loss proxy of \( m \) in the original loss function \( \mathcal{L}(m) \) at \( m_t \):

\begin{equation*}
\mathcal{L}(m^t) + \langle \hat{\nabla} \mathcal{L}(m^t), m - m^t \rangle + \frac{1}{2\eta_1} \| m - m^t \|^2
\end{equation*}

where ${\eta_1}$ is the learning rate for $m$. Therefore, the original problem can be simplified to the following proximal optimization update:

\begin{equation*}
\arg\min_m \frac{1}{2} \left\| m - \bar{m} \right\|^2 + \eta_1 y \left\| m \right\|_{\lceil s \rceil,2}^2
\end{equation*}

where $\bar{m}=m^t-\eta_1\hat{\nabla}_m\mathcal{L}(m^t).$
We set $S(y^t,s^t,m) = {y^t\left\|m\right\|_{\lceil s^t \rceil,2}^2}$ and the solution to the proximal operator $\mathrm{Prox}_{\eta_1S(y^t,s^t,m)}(\bar{m})$ can be defined as follows:

\begin{equation*}
m_i^* = 
\begin{cases}
\bar{m}_i, & \mathrm{if}\,\bar{m}_i \geq \bar{m}_{\text{least-}\lceil s\rceil} \\
\frac{1}{1+2\eta_1y} \bar{m}_i, & \text{otherwise}
\end{cases}
\end{equation*}

Here, $m_i$ represents the $i$-th element of the mask and least-$j$ denotes the index of the element in $m$ with the $j$-th smallest norm.

Unlike previous optimization-based methods, we do not use reparameterization to force $m$ to polarize to 
0 and 1. Instead, we adopt a uniformly distributed normal mask, allowing $m$ to update freely within the range of 0 to 1. This approach continually decays $m$ during the proximal optimization update to achieve pruning. Meanwhile, $\eta_1$ as the decay rate, can be freely adjusted as a hyperparameter and does not need to match the learning rate of the mask during actual optimization. In this process, the mask itself acts as a scaling factor for the weights, and the masks corresponding to unpruned weights can also attain intermediate values between 0 and 1 during optimization, thereby scaling the entire weight group and fine-tuning the weights. Since the mask can be freely optimized and is not limited to 0 and 1, in the actual pruning process, it is necessary to fuse the masks that are not equal to 1 into the weights. For the Llama model, the MHA and FFN components respectively scale the $W_V$ weights and the $W_{gate}$, $W_{up}$ weights with the weight group to maintain weight uniformity before and after pruning:

\begin{equation*}
\hat{W}_{V}^{(l, j)} = W_{V}^{(l, j)} \cdot m_{head}^{(l, j)}
\end{equation*}
\begin{equation*}
\hat{W}_{gate}^{l} = W_{gate}^{l} \cdot m_{inter}^{l}
\end{equation*}
\begin{equation*}
\hat{W}_{up}^{l} = W_{up}^{l} \cdot m_{inter}^{l}
\end{equation*}

In the optimization objective, the gradient of $s$ is related to two terms. The second term's resource constraint is typically a differentiable function of $s$, allowing for direct computation of its gradient $\tilde{\nabla}_{\boldsymbol{s}} z\left(M(s)-M_{prune}\right)$. However, the first term's sparsity constraint is non-differentiable. Specifically, $s$ is a floating-point number representing the model's dimension value, in practice, the ceiling function should be used to determine the integer number of dimensions to be pruned for each layer. However, the ceiling function is non-differentiable. To address this issue, apply the Straight-through Estimator \cite{bengio2013estimatingpropagatinggradientsstochastic} to provide an approximate gradient during backpropagation, i.e.,$\frac{\tilde{\partial}\lceil s\rceil}{\tilde{\partial}s}=1.$

For the sparsity loss involving $\left\|m\right\|_{s,2}^2$, using $\left\|m\right\|_{s+1,2}^2-\left\|m\right\|_{s,2}^2$ as the approximate proxy value for the partial derivative of $\left\|m\right\|_{s,2}^2$ :

\begin{equation*}
\frac{\tilde{\partial}\left\| m \right\|_{s,2}^2}{\tilde{\partial}s} = m_{\text{least-min}\{\mathrm{Dim}(m),s+1\}}^2
\end{equation*}

where $\mathrm{Dim}(m)$ is the number of elements in $m$.The variables $y$ and $z$ are coefficients of the Lagrangian penalty terms, which need to be continuously increased during the optimization process to minimize their corresponding penalty terms, thereby ensuring that the optimization objectives for $s$ and $m$ meet the desired sparsity goals. Specifically, we use gradient ascent to update them as follows:

\begin{equation*}
y^{t+1} = y^t + \eta_3 \| m^{t+1} \|_{\left\lceil s^{t+1} \right\rceil,2}^2
\end{equation*}
\begin{equation*}
z^{t+1} = \max(0, z^t + \eta_4 (M(s^{t+1}) - M_{prune}))
\end{equation*}

\subsection{Optimization with LoRA and Distillation}
LoRA \cite{hu2021lora} has been widely demonstrated to be efficient in fine-tuning LLMs. To effectively update weights during the optimization of the mask and achieve enhanced performance, similar to Compresso \cite{guo2023compresso} and NutePrune \cite{li2024nuteprune}, we introduce the LoRA module during optimization:
$$
W^{\prime}={W}+\Delta W=W+BA
$$

where $B \in  {\mathbb{R}}^{d \times  r},A \in  {\mathbb{R}}^{r \times  k}\;$ and $\;r \ll  \min \left( {d,k}\right)$. During the training process, the model's original weights $W$ are frozen, and only the parameters in the low-rank matrices $A$ and $B$ are trained.

Regarding loss functions, we introduce a distillation loss similar to those in CoFi \cite{xia2022structured} and NutePrune \cite{li2024nuteprune}. Specifically, $\mathcal{L}_{KL}$
denotes the Kullback-Leibler (KL) divergence between the probability distributions ${\mathbf{p}}_{t}$ and ${\mathbf{p}}_{s}$ of the output from the teacher model before pruning and the student model after pruning, respectively. Additionally, $\mathcal{L}_{layer}$ represents the sum of mean squared errors (MSE) of the hidden representations $h_s^l$ and $h_t^l$ cross the intermediate layers of the teacher and student models:

$$
{\mathcal{L}}_{\text{KL}} = {D}_{\mathrm{{KL}}}\left( {{\mathbf{p}}_{s}\parallel {\mathbf{p}}_{t}}\right)
$$
$$
{\mathcal{L}}_{\text{layer }} = \mathop{\sum }\limits_{l=1}^{L}\operatorname{MSE}(h_s^l,h_t^l)
$$
The coefficient of the two losses is controlled by the hyperparameter $\alpha$, and the final loss is formulated as:
$$
\mathcal{L}_{distill}=\mathcal{L}_{KL}+\alpha * \mathcal{L}_{layer}
$$

\section{Experiments}

\subsection{Setup}

\paragraph{Model}

To validate the effectiveness and generalizability of our method, we conducted experiments on several models, including the Llama-1 \cite{touvron2023llama} and Llama-2 \cite{touvron2023llama2} families, encompassing 7B and 13B configurations.

\paragraph{Implementation Details}

We sampled 20,000 data instances, each consisting of 512 tokens, from the C4 dataset \cite{raffel2020exploring} to serve as training data for mask optimization using the AdamW optimizer. The learning rate was set to 1e-2 for the mask parameters and 1e-3 for the LoRA parameters, with a batch size of 16. The pruning process was conducted over 7 epochs, during which the sparsity of each dimension incrementally increased until the target total sparsity was achieved. For the masks corresponding to the intermediate dimensions of FFN, proximal gradient updates were not performed at every step. Instead, updates occurred every $t$ iteration, while regular gradient descent was conducted at other times. The parameter $t$ decreased linearly from 10 to 1 throughout the optimization process. The target model was obtained directly after pruning without post-fine-tuning. All experiments were executed on a single NVIDIA A100 GPU.

\paragraph{Evaluation}

We initially assessed the pruned model's language modeling capability by measuring its perplexity on the Wikitext \cite{merity2016pointer} dataset. Furthermore, to ensure consistency with previous approaches, we conducted a comprehensive evaluation of the model's zero-shot capabilities using the lm-evaluation-harness \cite{eval-harness}. This evaluation encompassed zero-shot tasks on common sense reasoning datasets, including BoolQ \cite{clark2019boolq}, PIQA \cite{bisk2020piqa}, HellaSwag \cite{zellers2019hellaswag}, WinoGrande \cite{sakaguchi2020winogrande}, ARC-easy \cite{clark2018think}, ARC-challenge \cite{clark2018think} and OpenbookQA \cite{mihaylov2018can}.

\paragraph{Baseline}

We compared our method with the widely used LLM-Pruner \cite{ma2023llm}. Since our approach is based on optimization for structured pruning, we also evaluated it against methods with similar objectives, such as Compresso \cite{guo2023compresso} and NutePruner \cite{li2024nuteprune}. Although we adopted the same experimental settings as these methods, MaskPrune is disadvantaged due to the inherent layer alignment characteristic of our approach, especially when compared to methods like NutePrune \cite{li2024nuteprune}, which utilize different inter-layer structures. The NutePrune codebase provides a layer-uniform loss function similar to SheardLlama \cite{xia2023sheared}. We employed this configuration to reproduce NutePrune-uniform, thereby ensuring a more equitable comparison. However, due to the intrinsic properties of this loss function, complete alignment could not be achieved within the same 7 epochs; while achieving full alignment required up to 36 epochs. For optimization-based methods, this extended time cost is impractical. Therefore, we increased the coefficient of the sparsity loss term in NutePrune's loss function to attain a balanced state as effectively as possible within 7 epochs.

\subsection{Main Results}

\begin{table*}
  \centering
  \resizebox{\linewidth}{!}{
  \begin{tabular}{@{}lccccccccccl@{}}
    \hline
    \textbf{Ratio} & \textbf{Method} & \textbf{WikiText2$\downarrow$} & \textbf{BoolQ} & \textbf{PIQA} & \textbf{HellaSwag} & \textbf{Winogrande} & \textbf{ARC-e} & \textbf{ARC-c} & \textbf{OBQA} & \textbf{Avg.$\uparrow$} \\ 
    \hline
    0\%  & LLaMA-7B & 5.68 & 73.18 & 78.35 & 72.99 & 67.01 & 67.45 & 41.38 & 42.40 & 66.39 \\ 
    \hline
    20\% 
    & LLM-Pruner & 9.96 & 59.39 & 75.57 & 65.34 & 61.33 & 59.18 & 37.12 & 39.80 & 59.01 \\ 
    & Compresso & 10.38 & \textbf{73.64} & 75.08 & 64.77 & \textbf{67.72} & 66.12 & 37.54 & \textbf{40.40} & 60.75 \\
    & NutePrune-uniform & 8.74 & 71.01 & \textbf{76.55} & 67.97 & 65.75 & 68.10 & 36.60 & 38.20 & 60.59 \\
    & \textbf{MaskPrune} & \textbf{7.77}  & 68.50 & 76.17 & \textbf{69.84} & 66.30 & \textbf{68.56} & \textbf{39.68} & 39.20 & \textbf{61.17} \\
    \hline
    25\%
    & NutePrune-uniform & 9.48 & 66.85 & \textbf{75.52} & 65.92 & 61.33 & \textbf{66.29} & 34.64 & 36.00 & 58.07 \\
    & \textbf{MaskPrune} & \textbf{8.70} & \textbf{67.77} & 75.41 & \textbf{66.54} & \textbf{61.40} & 64.90 & \textbf{35.75} & \textbf{37.40} & \textbf{58.45} \\
    \hline
    50\% 
    & LLM-Pruner & 98.10 & 52.32 & 59.63 & 35.64 & 53.20 & 33.50 & 27.22 & 33.40 & 40.94 \\ 
    & Compresso & 59.73 & 60.09 & 66.70 & 39.31 & 51.93 & 48.82 & \textbf{27.82} & 33.40 & 46.87 \\
    & NutePrune-uniform & 20.69 & 62.84 & 68.50 & \textbf{50.76} & 54.22 & 48.99 & 26.96 & \textbf{33.60} & 49.41 \\
    & \textbf{MaskPrune} & \textbf{19.33} & \textbf{63.39} & \textbf{70.73} & 49.96 & \textbf{55.56} & \textbf{50.76} & 25.68 & 33.40  & \textbf{49.92} \\
    \hline
  \end{tabular}
  }
  \caption{\label{Llama-7B}
    Performance on zero-shot tasks and perplexity on the Wikitext2 dataset for the compressed Llama-7B model. NutePrune-uniform denotes the version of NutePrune with uniform inter-layer structures.
  }
\end{table*}

\paragraph{Zero-Shot Tasks}

For the Llama-7B baseline model, we applied pruning to generate models with three sparsity levels: 50\%, 25\%, and 20\%. As shown in Table \ref{Llama-7B}, our method outperforms Compresso \cite{guo2023compresso}, which employs uneven inter-layer sparsity, in terms of both perplexity and common sense reasoning tasks, demonstrating superior performance under challenging conditions. Compared to the layer-uniform version of NutePrune \cite{li2024nuteprune}, our method maintains higher model capabilities at each sparsity level, showing improvements of 0.58\%, 0.38\%, and 0.51\%, respectively. These results indicate that our method can consistently identify appropriate modules for pruning during optimization and effectively scale the remaining modules to preserve model performance.

Our method also demonstrates superior capabilities for larger models, such as Llama-13B and Llama-2-13B. Specifically, for Llama-13B and Llama-2-13B at a 20\% sparsity level, our method maintains 94\% and 95\% of the original model's zero-shot task capabilities, with no significant decrease in perplexity. 

\paragraph{Uniformity}

\begin{figure}
    \centering
    \includegraphics[width=\linewidth]{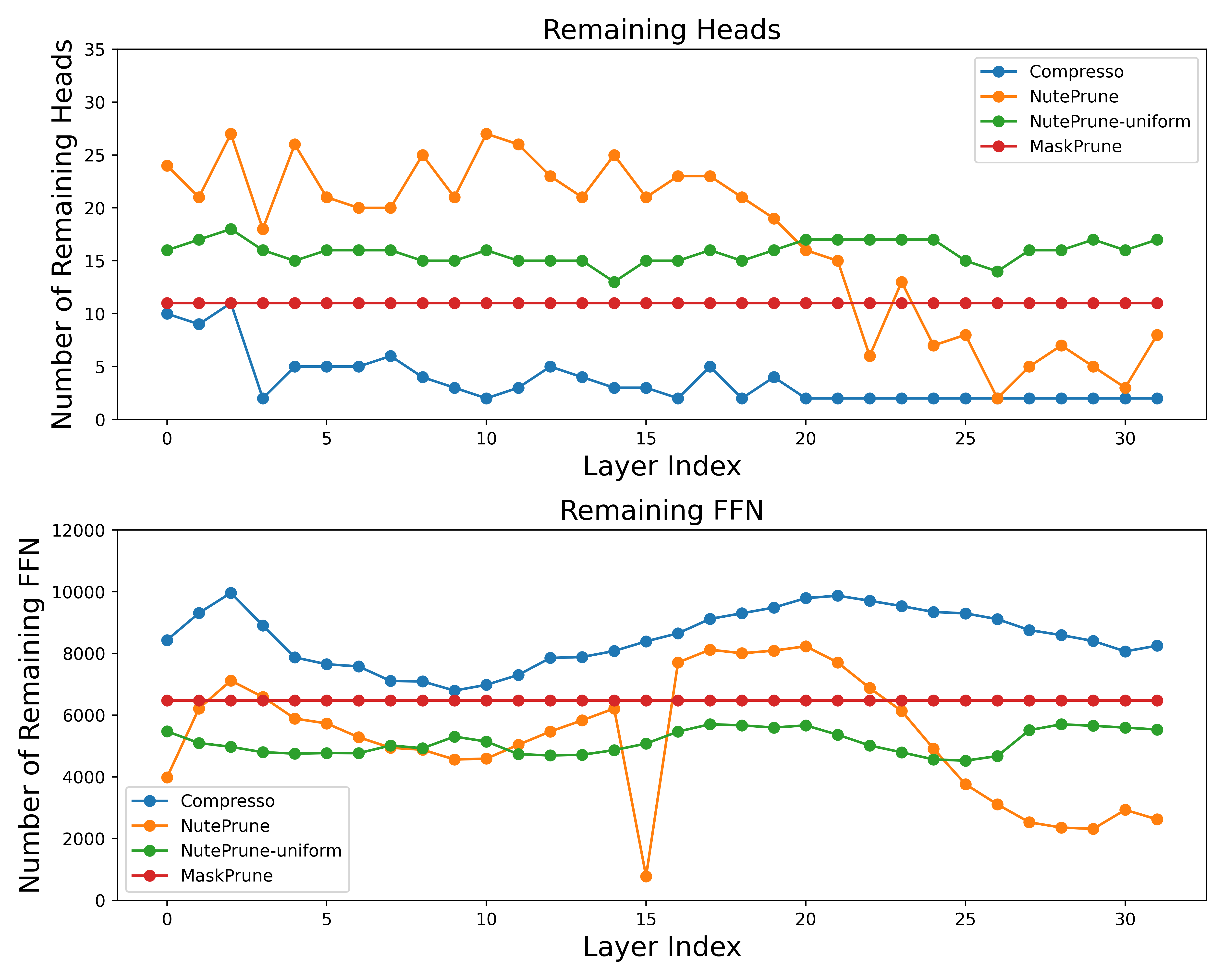}
    \caption{The number of heads and FFN intermediate dimensions retained after pruning Llama-7B to a sparsity of 50\% }
    \label{fig:uniform}
\end{figure}

As illustrated in Figure~\ref{fig:uniform}, our method continuously regularizes the sparsity across the model's layers to maintain uniformity throughout the training process, ultimately achieving a completely uniform structure. In contrast, Compresso and NutePrune permit unrestricted learning of masks during training, resulting in significant disparities in sparsity between layers, especially at higher sparsity levels. For instance, at 50\% sparsity of Llama-7B, some layers retain only the 2 attention heads while others retain the 27 attention heads. This irregular sparsity distribution poses greater challenges for the model during training and post-processing. The layer-uniform version of NutePrune essentially maintains uniformity across layers but struggles to achieve complete uniformity, ultimately having to compromise by reducing sparsity or performance during the actual pruning stage. In contrast, our method inherently maintains a uniform inter-layer structure, consistently preserving uniformity after the training process concludes.


\begin{figure}
    \centering
    \includegraphics[width=\linewidth]{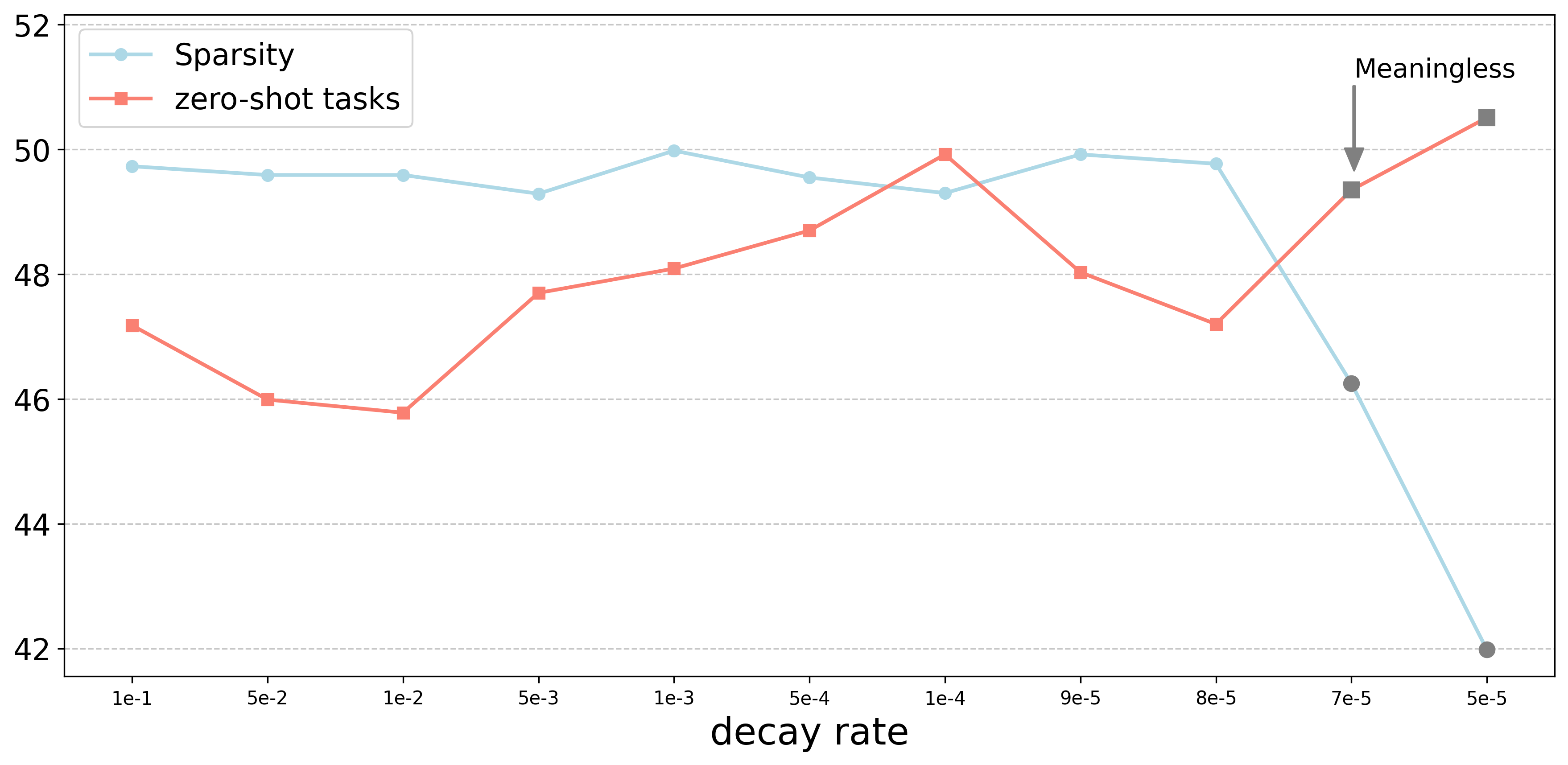}
    \caption{Zero-shot performance and actual sparsity of the Llama-7B model at 50\% sparsity under different decay rates.}
    \label{fig:Decay Rate}
\end{figure}

\begin{figure}
    \centering
    \includegraphics[width=1\linewidth]{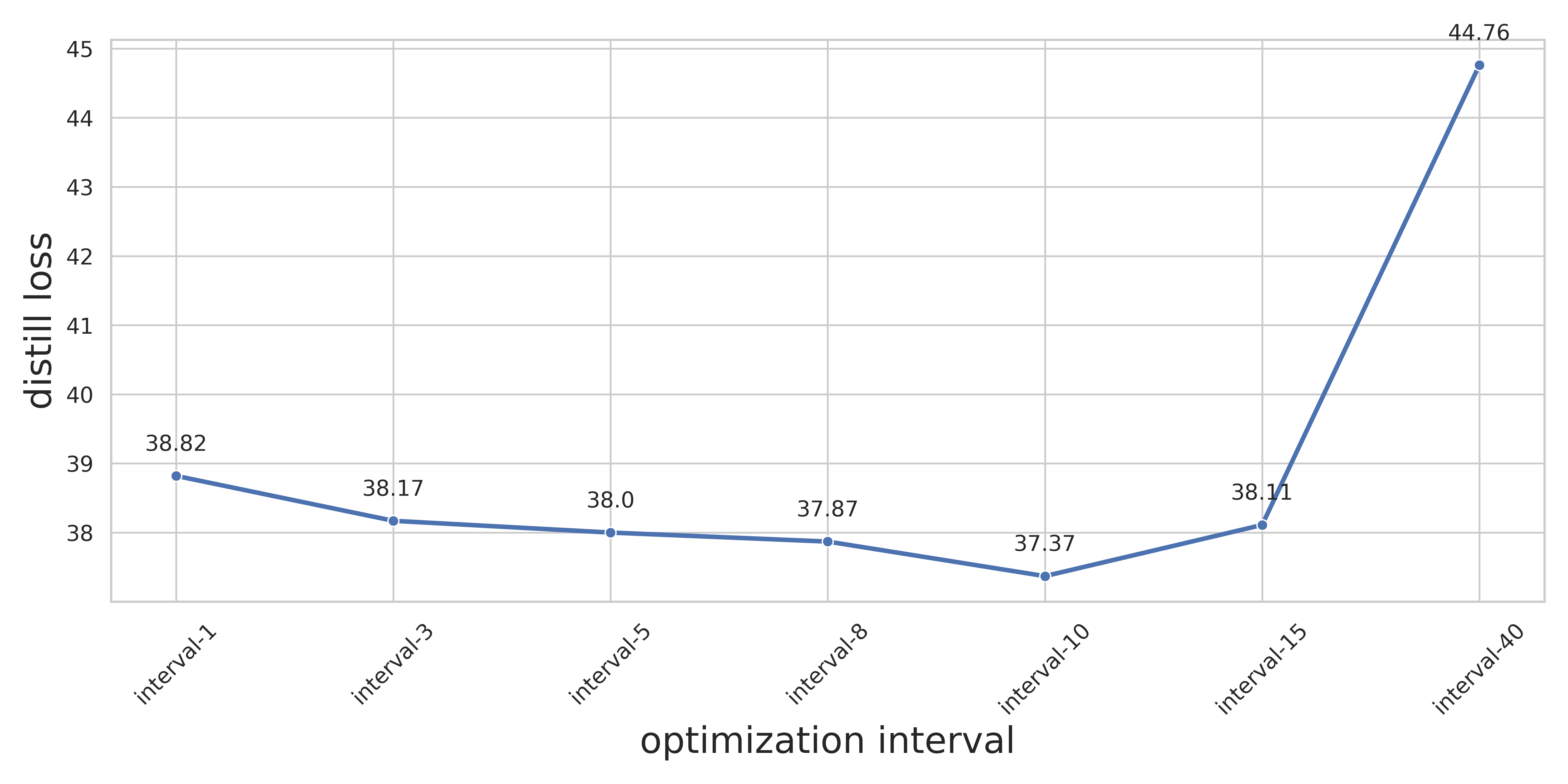}
    \caption{Training loss under different optimization intervals}
    \label{fig:Optimization Intervals Loss}
\end{figure}

\begin{figure}[ht]
    \centering
    \includegraphics[width=1\linewidth]{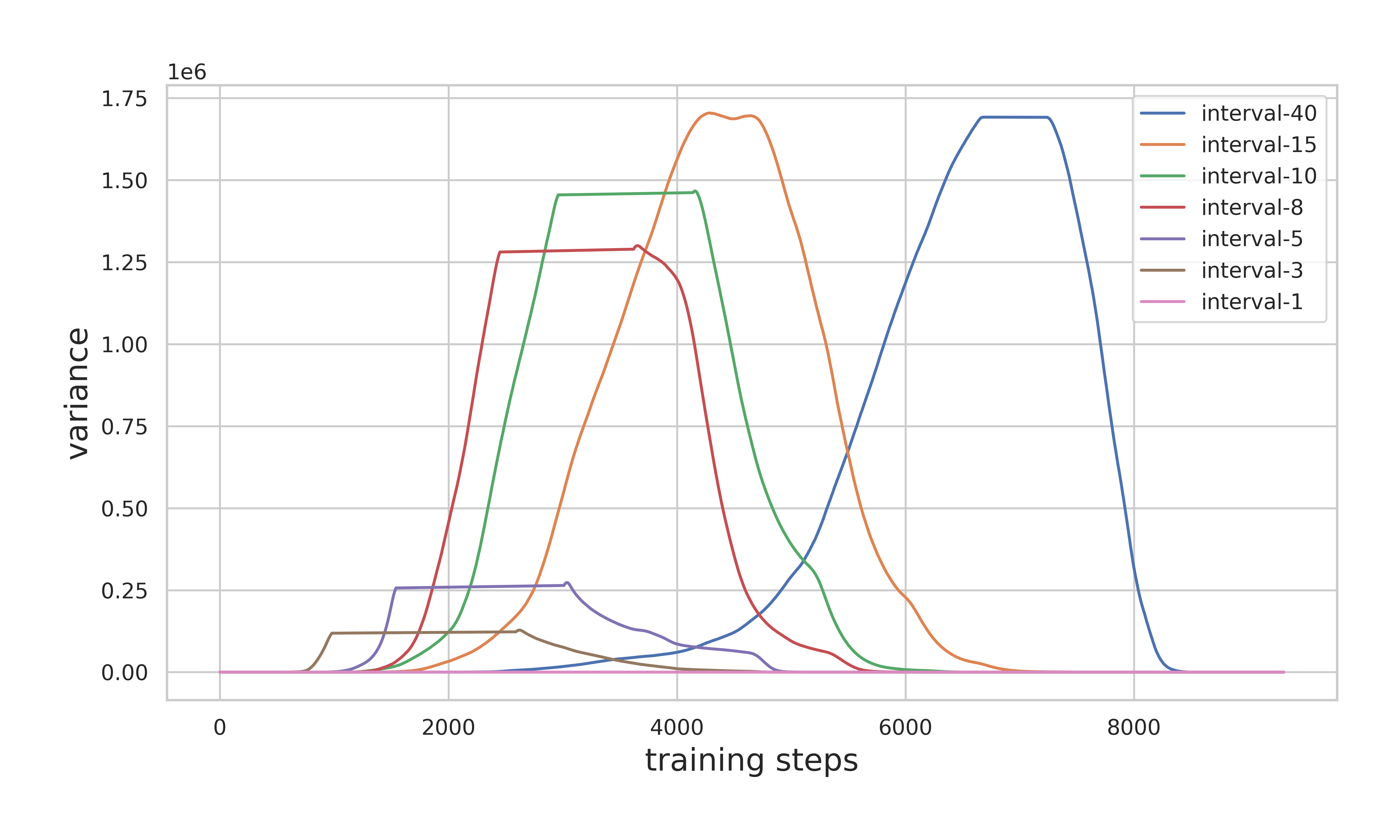}
    \caption{Variance changes in the intermediate dimensions of each FFN layer during the pruning process under different optimization intervals.}
    \label{fig:Optimization Intervals Variance}
\end{figure}

\begin{table*}
  \centering
  \resizebox{\linewidth}{!}{
  \begin{tabular}{@{}lccccccccccl@{}}
    \hline
    \textbf{Ratio} & \textbf{Method} & \textbf{WikiText2$\downarrow$} & \textbf{BoolQ} & \textbf{PIQA} & \textbf{HellaSwag} & \textbf{Winogrande} & \textbf{ARC-e} & \textbf{ARC-c}
    & \textbf{OBQA} & \textbf{Avg.$\uparrow$} \\ 
    \hline
    0\%  & LLaMA-13B & 5.62 & 77.68 & 79.49 & 77.01 & 72.69 & 76.68 & 45.99 & 44.00 & 67.64 \\
    \hline
    20\% 
        & NutePrune-uniform & \textbf{7.59} & \textbf{73.91} & \textbf{77.75} & 71.68 & \textbf{68.75} & 71.51 & 39.76 & 40.40 & 63.39 \\
        & \textbf{MaskPrune} & 8.25 & 73.00 & 77.15 & \textbf{72.21} & 68.51 & \textbf{72.14} & \textbf{41.21} & \textbf{41.00} & \textbf{63.60} \\
    \hline
    0\%  & LLaMA-2-13B & 5.30 & 82.05 & 77.86 & 77.22 & 72.77 & 78.37 & 47.18 & 44.80 & 68.83 \\
    \hline
    20 \%
        & NutePrune-uniform & 7.40 & 76.24 & 76.55 & 71.53 & 67.48 & 71.76 & 39.93 & 41.00 & 63.52 \\
        & \textbf{MaskPrune} & \textbf{6.64} & \textbf{76.70} & \textbf{77.69} & \textbf{73.06} & \textbf{70.80} & \textbf{74.16} & \textbf{43.26} & \textbf{41.60} & \textbf{65.32} \\
    \hline
  \end{tabular}
  }
  \caption{\label{Llama-13B}
    Performance on zero-shot tasks and perplexity on the Wikitext2 dataset for the compressed Llama-13B and Llama-2-13B model.
  }
\end{table*}

\subsection{Analysis}

\paragraph{Type of Mask}

\begin{table}
  \small
  \centering
  \begin{tabular}{@{}lcl@{}}
    \hline
    \textbf{Model} & \textbf{Type of mask} & \textbf{Avg.$\uparrow$} \\ 
    \hline
    & \textbf{Uniform (Ours)} & \textbf{49.92} \\
    LLaMA-7B & $L_0$ \cite{louizos2018learning} & 45.36 \\
    & Polarization \cite{guo2021gdp} & 48.29 \\
\hline
  \end{tabular}
  \caption{\label{Type of Mask}
    Impact of different types of masks on model compression performance.
  }
\end{table}

We explored the impact of different mask types on our method by selecting three representative mask distributions: (1) a standard uniform distribution mask, which can be freely optimized within the range of [0,1]; (2) a standard $L_0$ regularization mask \cite{louizos2018learning}, which drives the mask values to 0 and 1 through reparameterization; and (3) a differentiable polarizing mask \cite{guo2021gdp}. As shown in Table \ref{Type of Mask}, the uniform distribution mask performs the best. We speculate that this is because decaying the mask essentially involves a smooth transition from 1 to 0. During intermediate values, it is equivalent to scaling the model's weights, allowing the model to transition more effectively from dense to sparse. The steep distribution of the $L_0$ regularization mask conceals this advantage, preventing the mask from correctly attaining intermediate states. The differentiable polarizing mask gradually becomes steeper during training, and this changing distribution also alters the scaled weights. Additionally, our method naturally contracts to 0, eliminating the need to approximate the mask distribution using other distributions.

\paragraph{Decay Rate}

We also investigated the impact of different mask decay rates on the final model performance. As shown in Figure \ref{fig:Decay Rate}, experiments indicate that a decay rate that is too high causes masks with initially small values to quickly decay to 0, preventing the adjustment of incorrectly pruned weights. Conversely, a decay rate that is too low fails to counteract the mask's inherent optimization trend, preventing the mask from decaying to minimal values and achieving the desired sparsity and layer uniformity.

\paragraph{Optimization Interval}

In our method, we do not perform proximal gradient updates on the mask values at every iteration step. This approach can cause some masks to erroneously decay to zero rapidly and become irrecoverable in subsequent optimization processes, leading to the unintended pruning of weights and a significant decline in model performance. This phenomenon is particularly evident when pruning the intermediate dimensions of the FFN. To prevent this, during regular iteration steps, we only perform standard gradient descent on the masks and conduct proximal gradient updates at specified intervals to ensure that the masks achieve the target sparsity. The optimization interval linearly decreases from 10 to 1 throughout the optimization process, allowing the model to eventually converge to the desired sparsity level. We investigate the impact of the hyperparameter optimization interval on model pruning performance. As shown in Figure \ref{fig:Optimization Intervals Loss} and Figure \ref{fig:Optimization Intervals Variance}, we present the final training loss and the variance of the FFN dimensions across layers under different optimization intervals, respectively. The latter reflects the uniformity across model layers and typically exhibits a trend of initially increasing and then decreasing. It can be observed that when the optimization interval is set to 10, the training loss reaches its minimum. When the interval is smaller than this value, the model converges rapidly in the early stages of training, causing some masks to erroneously decay to zero and remain irrecoverable. Conversely, when the interval is larger than 10, the masks only begin to decay to zero as the optimization interval gradually decreases in the later stages, thereby wasting the early pruning process.




\section{Conclusion}
This paper introduces a structured pruning method for Large Language Models (LLMs) by formulating model pruning as a minimax optimization problem. During optimization, the pruning mask and the model's target dimensions are trained simultaneously, resulting in a pruned model with uniform inter-layer structures. We evaluated models with varying sparsity levels and sizes across multiple benchmarks, and the results consistently demonstrate the superiority of our method. Our approach also investigates different strategies for performing optimization-based structured pruning on LLMs, providing new insights into the compression of these models.

\section{Limitations}
The efficacy and effectiveness of mask learning-based pruning is highly dependent on the quality of the training or calibration data. Determining whether the chosen data is optimal and developing various strategies for selecting superior datasets remain significant challenges that warrant further investigation. 


\bibliography{custom}
\newpage
\appendix

\section{Mask Distribution}

\begin{figure}[ht]
    \centering
    \includegraphics[width=0.8\linewidth]{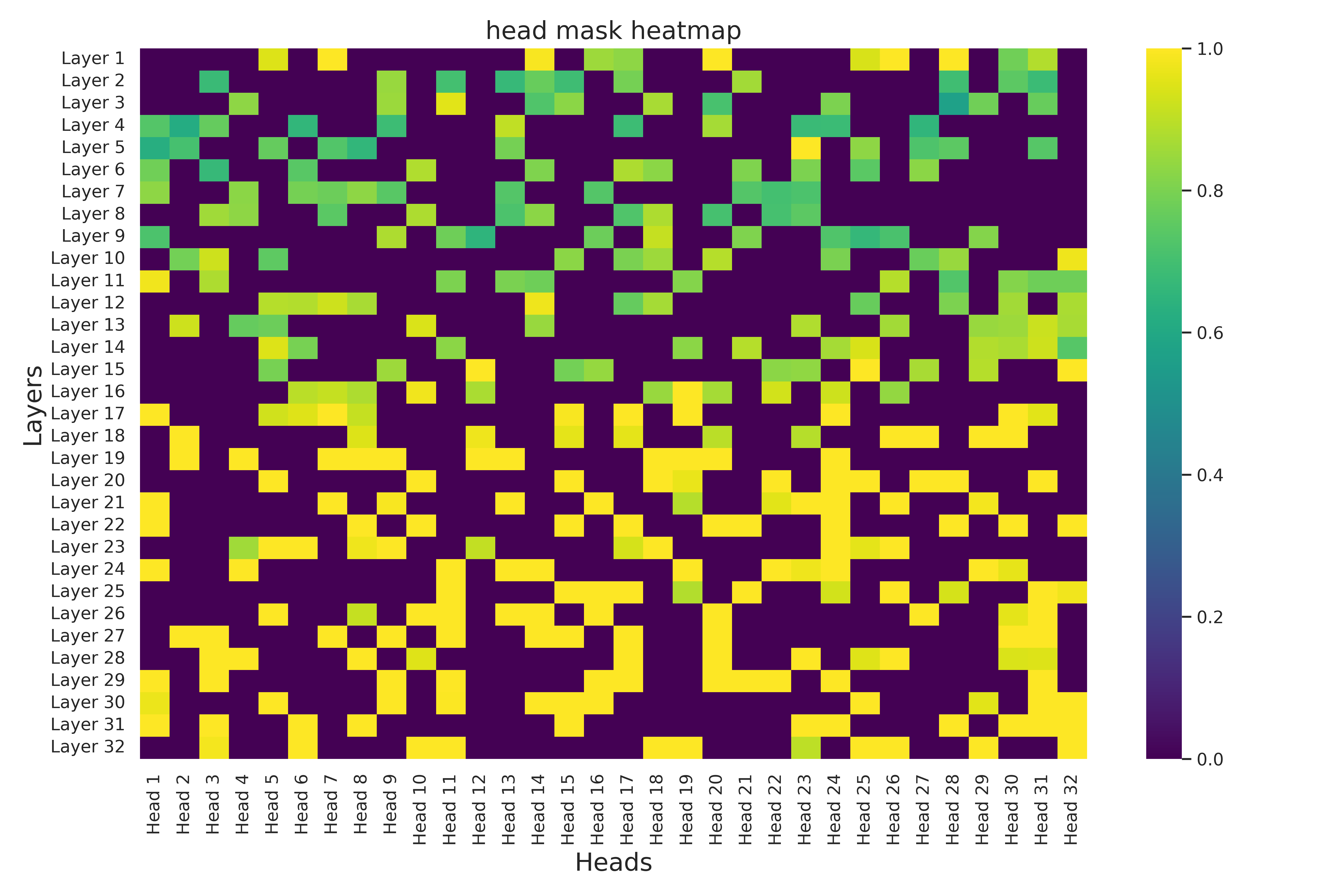}
    \caption{Heatmap of the value distribution of attention head pruning masks}
    \label{fig:head_z_heatmap}
\end{figure}

\begin{figure}[ht]
    \centering
    \includegraphics[width=\linewidth]{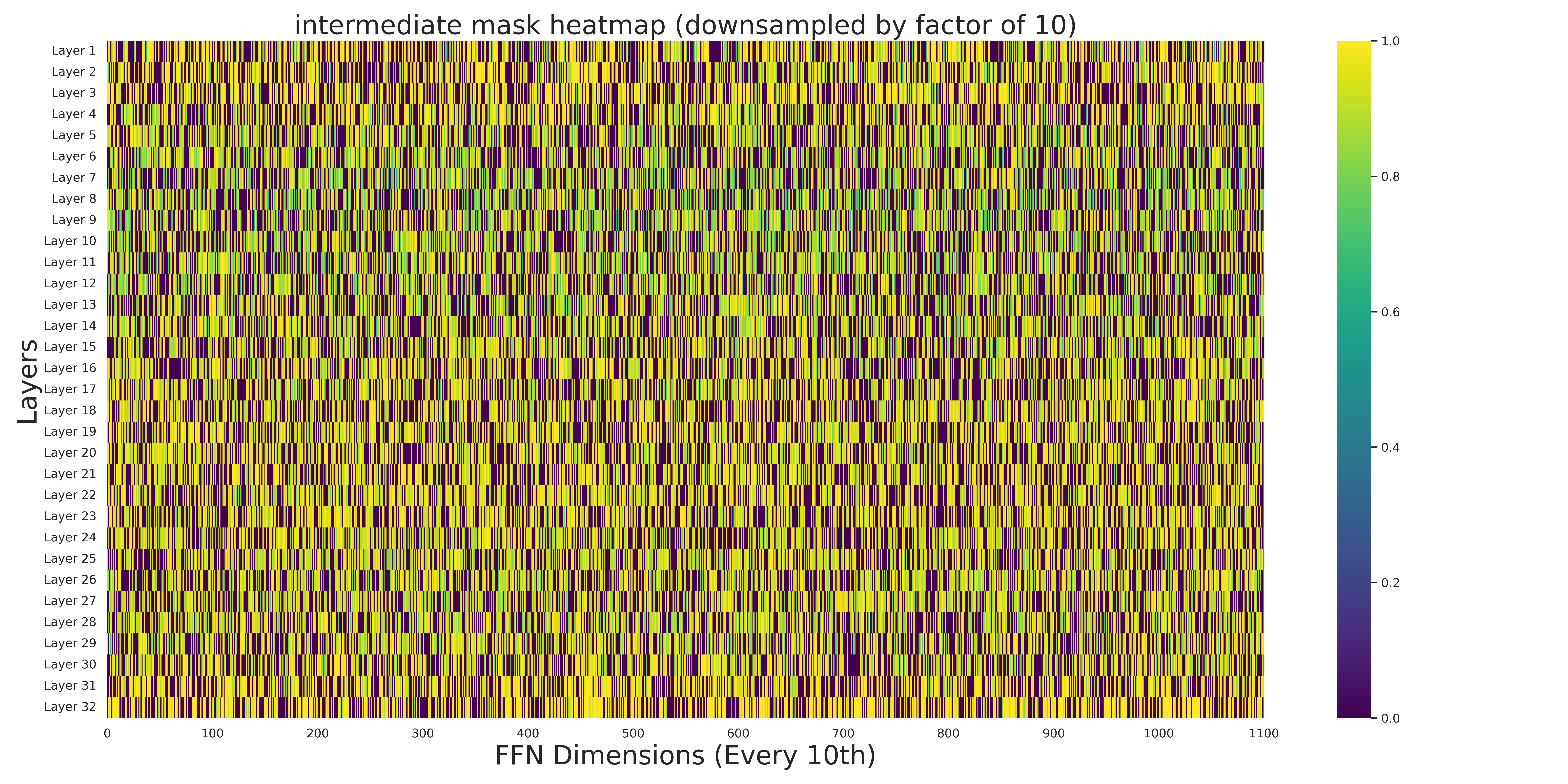}
    \caption{Heatmap of the value distribution of FFN intermediate pruning masks}
    \label{fig:intermediate_z_heatmap_downsampled}
\end{figure}

To investigate the effects of mask values across different layers, we analyzed spatial patterns through heatmap visualizations. Figure \ref{fig:head_z_heatmap} reveals distinct layer-wise patterns in the multi-head attention pruning masks. The first half of the model's layers show mask values predominantly clustered around 0.8, suggesting these layers employ parameter scaling to compensate for pruning-induced performance degradation. Conversely, the latter layers maintain values near 1, indicating lower pruning sensitivity where direct pruning minimally impacts model performance. Similar layer-specific patterns emerge in the FFN layers, as shown in Figure \ref{fig:intermediate_z_heatmap_downsampled}, where intermediate masks demonstrate varying distribution trends across layers, with some maintaining unpruned masks significantly below 1 while others approach unity.

\begin{figure}[ht]
    \centering
    \includegraphics[width=0.8\linewidth]{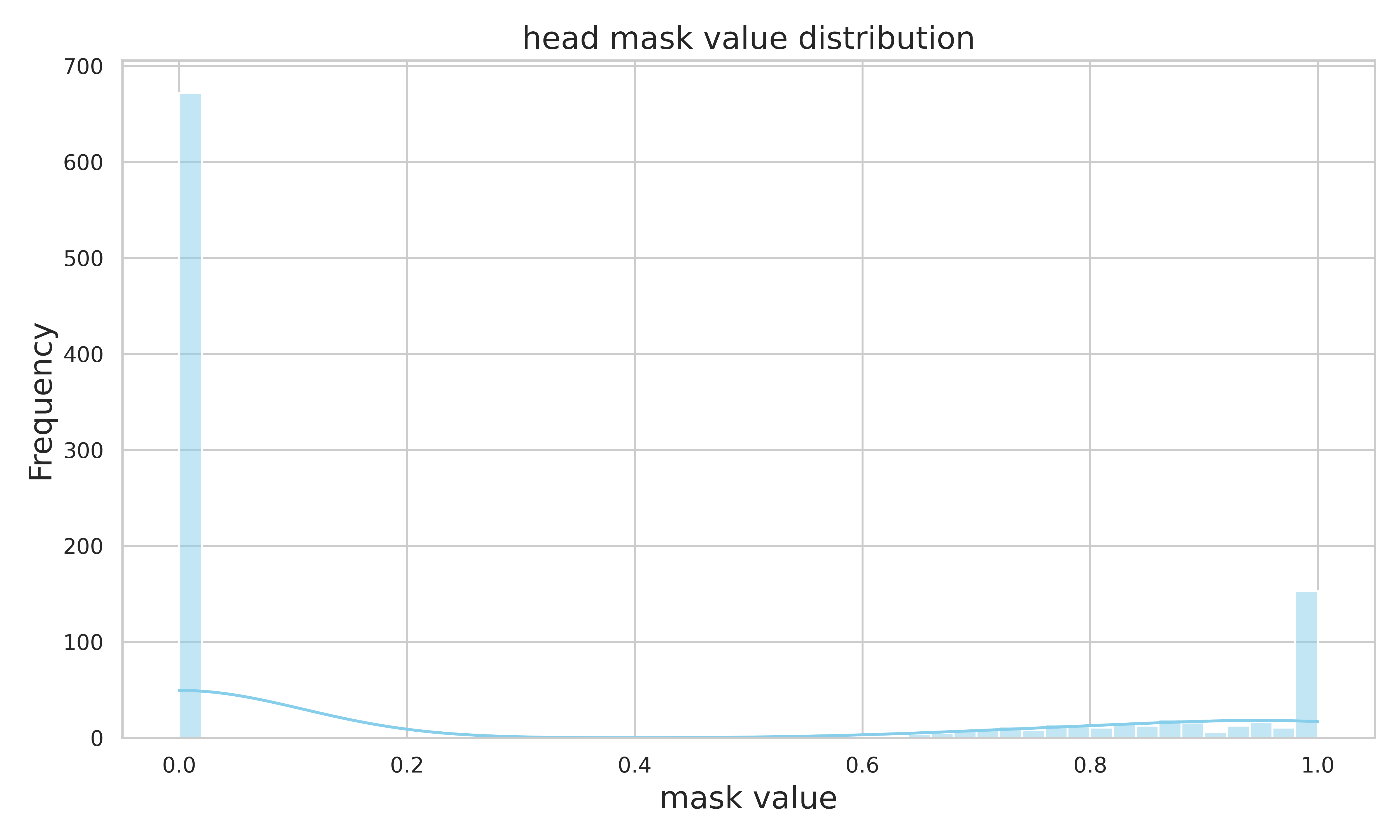}
    \caption{frequency distribution of attention
head pruning masks}
    \label{fig:head_z_distribution}
\end{figure}

\begin{figure}[ht]
    \centering
    \includegraphics[width=0.8\linewidth]{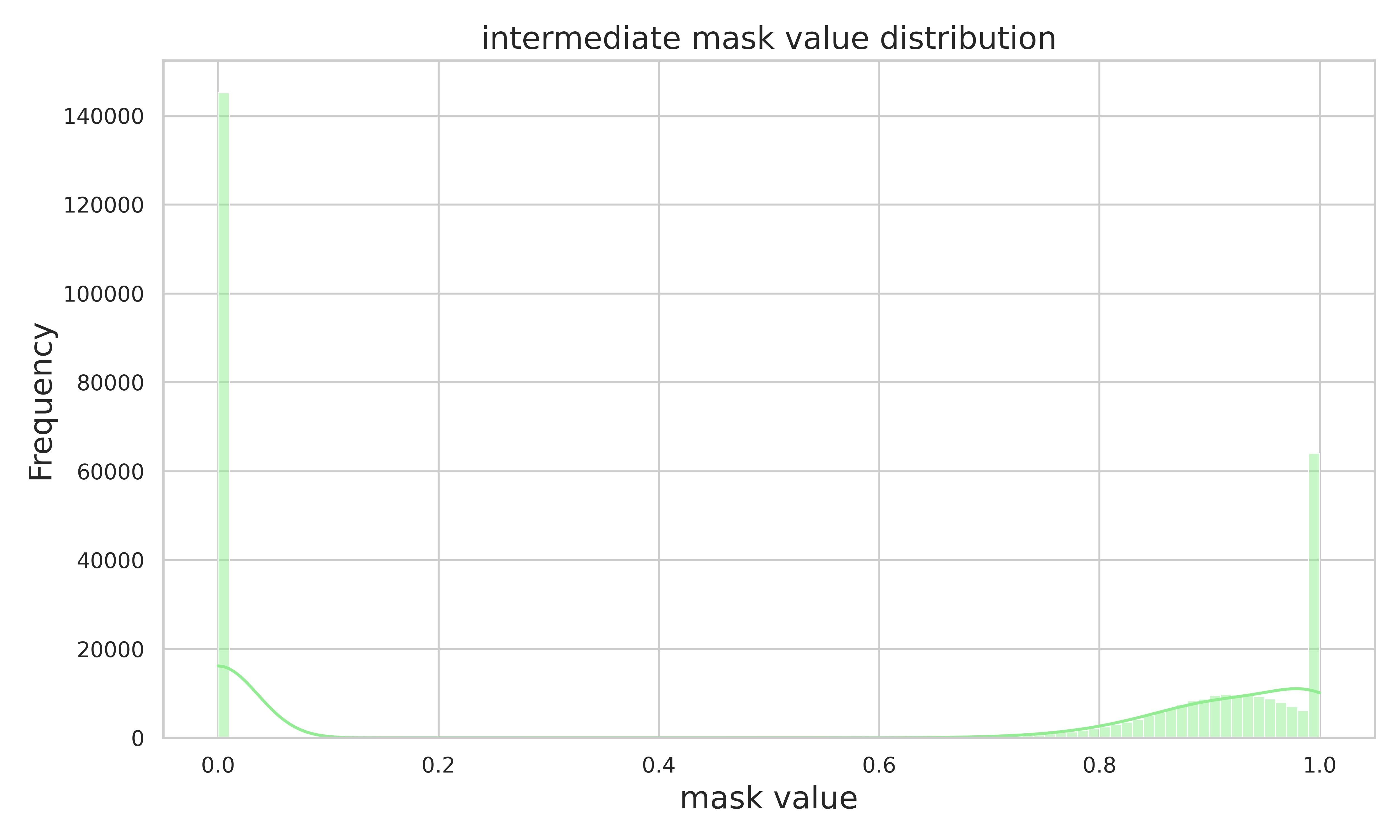}
    \caption{frequency distribution of FFN intermediate pruning masks}
    \label{fig:intermediate_z_distribution}
\end{figure}

The frequency distributions of mask values are presented in Figures \ref{fig:head_z_distribution} and \ref{fig:intermediate_z_distribution} for multi-head attention and FFN layers respectively. Both distributions exhibit bimodal characteristics with majority values clustered at the extremes (0 and 1), while maintaining a significant proportion of intermediate values. The multi-head attention masks show progressive value shifts across layers, with smaller values dominating early layers and values approaching 1 in deeper layers, confirming the increased pruning sensitivity in initial layers. Similarly, FFN intermediate masks display layer-dependent distribution patterns, with varying mean values across different network depths.

\section{Algorithm}
The pseudocode of the algorithm used in this paper is shown in Algorithm \ref{alg:main}.

\begin{algorithm2e}[htbp]
\small
\SetAlgoLined
\KwIn{Pruned Size $M_{\text{prune}}$, learning rates $\eta_1, \eta_2, \eta_3, \eta_4$, number of total iterations $\tau$.}
\KwResult{mask $m^*$}
Initialize $t=1$;\\
Initialize $m^1=1$;\\
\While{$t\leq\tau$}{
    $ m^{t+1} = \textrm{Prox}_{\eta_1 S(y^t, s^t, m^t)}(m^t - \eta_1 \hat{{\nabla}}_{m} \mathcal{L}( m^t));$ \\
    $s^{t+1} = s^t - \eta_2 \left(\tilde{\nabla}_s S(y^t, s^t, m^{t+1}) + \tilde{\nabla}_s ( z^t (M(s^t) - M_{\text{prune}}))\right);$ \\
    $y^{t+1} = {y}^{t} + \eta_3 \|m^{t+1}\|_{\lceil {s}^{t+1} \rceil, 2}^2;$ \\ 
    $z^{t+1} = \max(0, z^t + \eta_4 (M(s^{t+1}) - M_{\text{prune}}));$ 
}
$m^* = m^{\tau}$.
\caption{Gradient-based algorithm.}
\label{alg:main}
\end{algorithm2e}

\end{document}